\documentclass[letterpaper, 10 pt, conference]{ieeeconf}  
\IEEEoverridecommandlockouts                           
\usepackage{amsmath,amssymb}
\usepackage{graphicx}
\usepackage{booktabs}
\usepackage{algorithm}
\usepackage{algorithmic}
\usepackage{xcolor}
\usepackage{cite}
\usepackage{threeparttable}

\newcommand{\method}{TRIAGE}

\newcommand{\pp}{\,pp}

\title{\LARGE \bf
TRIAGE: Efficient Vision-Language-Action Inference Needs \\ Both Reuse and Pruning
}

\author{Sooyoung Kwon$^{1}$, Carlee Joe-Wong$^{2,*}$, and Kyunghan Lee$^{1,*}$%
\thanks{$^{1}$Dept. of ECE, Seoul National University.
        {\tt\small \{sooyoung7896, kyunghanlee\}@snu.ac.kr}}%
\thanks{$^{2}$Dept. of ECE, Carnegie Mellon University.
        {\tt\small cjoewong@andrew.cmu.edu}}%
\thanks{$^{*}$Corresponding authors.}%
}

\begin{document}

\maketitle
\thispagestyle{empty}
\pagestyle{empty}


\begin{abstract}
Vision-language-action (VLA) models are increasingly used for robot manipulation because they map camera images and a language instruction directly to actions. Robots that cannot be tethered to a workstation, such as mobile manipulators and humanoids, must run them on onboard embedded boards, where the prefill latency of the language backbone grows with the hundreds of visual tokens processed at every control step. To meet real-time limits, training-free accelerators recompute only a subset of the visual tokens and skip the rest, either by pruning them or by reusing their cached key-value (KV) states. The two options save nearly the same computation, yet existing methods apply only one of them to every skipped token. Through a token-level diagnostic, we show that which option keeps the action closer to full recomputation depends on the token's drift, the change in its shallow-layer value since its cache was computed. On both backbones tested, reusing tokens below a drift threshold was no worse than pruning them, whereas beyond it, reuse yields up to 4.4$\times$ the action error of pruning; the same diagnostic calibrates this threshold per backbone. Building on this finding, we propose \method, which deems a cache valid when its drift is below this threshold and it is not too old. \method\ recomputes a fixed budget of tokens, prioritizing important tokens with invalid caches, reuses every other token with a valid cache, and prunes the rest; prune-only and reuse-only methods are the two extremes of this partition. Because the budget fixes the number of recomputed tokens, it directly sets the prefill latency. On LIBERO and SimplerEnv, \method\ stays within 3\,pp of full recomputation in average success rate while cutting prefill latency on a Jetson AGX Orin by up to 48\%, and with only 7\% of visual tokens recomputed, it beats the strongest baseline by 24.4\,pp on LIBERO-Object.
\end{abstract}

\begin{figure}[!t]
  \centering
  \includegraphics[width=\columnwidth]{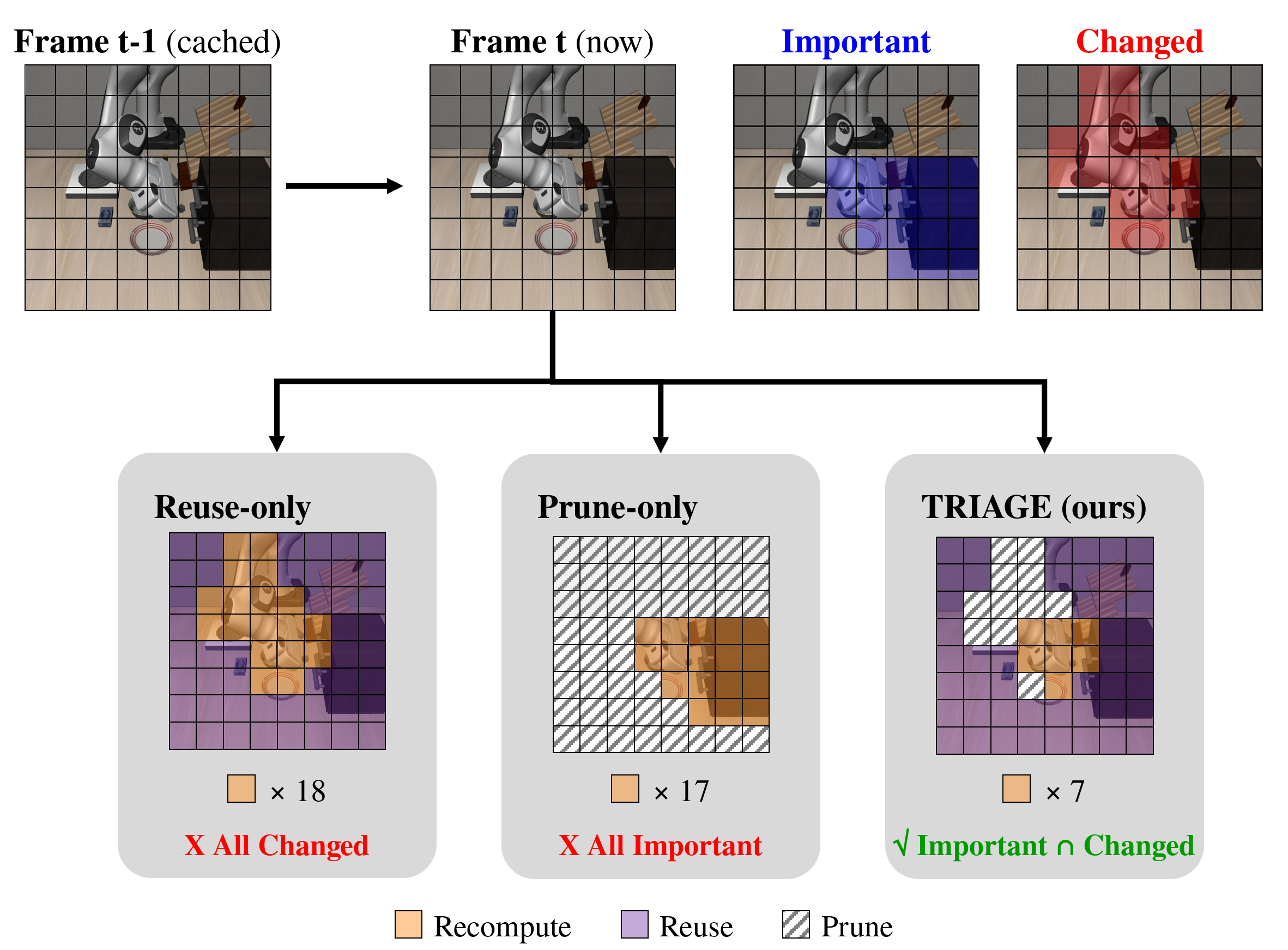}
  \caption{\textbf{Recompute only what neither reuse nor pruning can handle.} Blue: important tokens; red: tokens changed since frame $t-1$ (stale caches). Reuse-only recomputes every changed token (18) and prune-only every important token (17). \method\ recomputes only tokens that are both (7), reuses valid caches, and prunes the rest.}
  \label{fig:1}
\end{figure}

\section{INTRODUCTION}

\label{sec:intro}
Vision-Language-Action (VLA) models have emerged as a promising approach to general-purpose robotic manipulation~\cite{rt2,openvla,oft,pi0,pi05}. At every control step, a VLA model feeds hundreds of visual tokens, together with a language instruction, into a large language model (LLM) backbone that outputs actions. However, on embedded boards such as NVIDIA Jetson, the hardware typically mounted on robots, LLM inference is too slow for tasks that demand rapid reactions such as catching a moving object. Offloading inference to a remote server is not a dependable alternative either: end-to-end latency then depends on the network as much as on the server~\cite{vlaperf}, wireless links to mobile robots suffer from fluctuating latency and intermittent outages, and a single delayed or lost response stalls the control loop. On such hardware, the prefill stage of the LLM backbone is compute-bound, so its latency grows with the number of input tokens~\cite{onemore}; for models that decode all actions in a single forward pass, such as OpenVLA-OFT, prefill accounts for two thirds of each control step (Table~\ref{tab:libero}). In VLA models, visual tokens account for roughly 90\% of the input sequence, making them the main contributor to both computation and latency.

Training-free acceleration methods therefore mainly focus on reducing the number of visual tokens the backbone computes through token pruning or reuse. Token pruning~\cite{fastv,sparsevlm,vlapruner,specprune,adp,safe} estimates each token's importance at a shallow layer and permanently discards less important tokens, directly shortening the sequence processed by subsequent layers. Because pruning is irreversible, aggressive pruning ratios tend to remove task-relevant tokens, sharply degrading accuracy. Token reuse~\cite{vlacache,lac,gatedcache,tvcache} instead identifies tokens that remain largely unchanged from the previous frame and reuses their key-value (KV) caches. However, these caches were computed in the context of the previous frame; as the scene changes, they become stale. Repeatedly reusing those stale KV caches causes errors to accumulate and severely degrades accuracy.

We argue that applying a single operation uniformly to all tokens outside the recomputed set inevitably sacrifices accuracy or wastes computation (Fig.~\ref{fig:1}). Consider a token that is important but unchanged: pruning can only recompute it, wasting computation on information already in the cache, or discard it, losing information the model needs. Reuse, conversely, retains unimportant tokens that could have been discarded, and keeps serving cached states even after they have become stale. To our knowledge, no prior VLA acceleration method chooses between the two per token: every existing method either prunes or reuses all of the visual tokens it does not recompute (Section~\ref{sec:related}).

When, then, should a token that is not recomputed be reused, and when should it be pruned? Answering this requires a per-token signal that is available before the deep layers run, costs no extra computation, and predicts whether serving a cached state will distort the action. We use drift, the relative change in a token's shallow-layer value since its cache was last computed. Similar change measures have been used to choose which tokens to recompute~\cite{eventful,cacheblend}, but prior works have not studied whether such a signal can decide between reuse and pruning. We examine it with a token-level diagnostic (Section~\ref{sec:diagnostic}), a controlled experiment that handles the same tokens once by reuse and once by pruning and compares each resulting action with that of full recomputation. On both backbones we tested, reusing tokens below a drift threshold was no worse than pruning them, whereas as drift rises beyond it, reuse yields up to 4.4$\times$ the action error of pruning, regardless of token importance. The threshold depends on the backbone, and rerunning the diagnostic on a new backbone provides it.

This finding separates the roles of two per-token signals. Cache validity, which requires drift below the threshold and a bounded cache age, decides whether a token that is not recomputed should be reused or pruned. Importance, the attention a token receives from the action-decoding tokens, decides where the limited recomputation budget should go: to important tokens whose caches are no longer valid, for which neither reuse nor pruning is adequate.

We therefore propose a three-set structure that assigns each visual token to recomputation, reuse, or pruning based on its importance and cache validity. Within a given recomputation budget, important tokens with invalid caches are recomputed first; among the remaining tokens, those with valid caches are reused and the others are pruned.

Experiments on two VLA backbones and two benchmarks, with latency measured on a Jetson AGX Orin, show that \method\ achieves success rates comparable to state-of-the-art pruning and reuse methods at similar compute while reducing prefill latency by up to 48\%. Its advantage grows as the recomputation budget shrinks: on LIBERO-Object with OpenVLA, \method\ stays within 3\pp\ of full recomputation when recomputing only 11\% of the visual tokens, and at 7\% it outperforms the strongest baseline by 24.4\pp.

Our contributions are as follows:
\begin{itemize}
  \item A token-level diagnostic showing that whether a non-recomputed token should be reused or pruned depends on its drift, regardless of its importance, which makes both prune-only and reuse-only methods suboptimal.
  \item \method, a three-set structure that assigns each visual token to recomputation, reuse, or pruning, with prune-only and reuse-only methods as its two extremes.
  \item Experiments on two backbones and two benchmarks showing that \method\ stays within 3\pp\ of full recomputation while reducing prefill latency on a Jetson AGX Orin by up to 48\%.
\end{itemize}

\section{RELATED WORK}
\label{sec:related}
\textbf{Vision-language-action models.} VLA models adapt pretrained vision-language models to map camera images and a language instruction to robot actions, decoding them as discrete tokens~\cite{openvla}, as a continuous chunk in a single forward pass~\cite{oft}, or with a flow-matching action expert~\cite{pi0,pi05}. In all of them, visual tokens dominate the backbone's input (256 per step for OpenVLA and 512 for OpenVLA-OFT), so reducing their cost benefits every decoding style. Quantization~\cite{qvla}, layer skipping~\cite{mole,deer}, compact backbones~\cite{smolvla}, and asynchronous execution~\cite{smolvla,rtc,jetsonpi} are orthogonal and can be combined with our method.

\textbf{Visual token pruning.} VLM methods prune tokens that receive little shallow-layer attention~\cite{fastv,sparsevlm}, merge similar tokens~\cite{tome,vidtldr}, or keep compact subsets selected by attention or diversity~\cite{pyramiddrop,visionzip,divprune}. VLA methods add temporal and action cues: smoothed action-decode attention~\cite{vlapruner}, deep-layer attention from the previous step~\cite{specprune}, motion-adaptive ratios~\cite{adp}, forecast saliency~\cite{safe}, model scheduling~\cite{spvla}, or learned selection~\cite{lightvla}. FlashVLA~\cite{flashvla} and EfficientVLA~\cite{efficientvla} also reuse whole actions or action-head features, yet among visual tokens they too discard every token they do not keep, including tokens whose caches are still accurate.

\textbf{Temporal reuse.} Video models and serving systems avoid recomputing unchanged inputs: Eventful Transformers~\cite{eventful} and D\'ej\`a Vu~\cite{dejavu} reuse per-token computation across frames, AutoGaze~\cite{autogaze} removes redundant patches before the ViT, and CacheBlend~\cite{cacheblend}, H$_2$O~\cite{h2o}, and DyCoke~\cite{dycoke} reuse, evict, or merge KV entries. For VLAs, VLA-Cache~\cite{vlacache} reuses the KV caches of static patches, and later methods learn the reuse policy~\cite{lac}, gate it by action confidence~\cite{gatedcache}, select reuse heads and layers~\cite{tvcache}, or fuse past and current tokens~\cite{ttf}. In all of these VLA methods, every visual token that is not recomputed is reused, even after its cache has become stale.

\textbf{Our work.} Prune-only and reuse-only methods send every token outside the recomputed set to a single fate. \method\ instead decides per token among recomputation, reuse, and pruning: it reuses accurate caches that pruning would discard, retaining their information without computation, and prunes stale caches that reuse would keep. Both families are thus special cases of its partition.

\section{When Is a Cached Token Worth Keeping?}
\label{sec:diagnostic}

\begin{figure}[!t]
  \centering
  \includegraphics[width=\columnwidth]{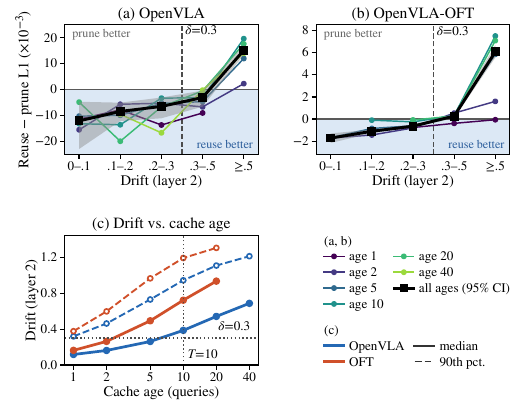}
  \caption{\textbf{Drift decides between reuse and pruning.} (a, b) Reuse $-$ prune action error $\Delta$ (Eq.~\eqref{eq:reuse_prune}, $\times 10^{-3}$) versus drift of the probed tokens on LIBERO-Spatial; blue region: reuse is better. Black: pooled over cache ages (95\% bootstrap CI); thin: individual ages. Reuse is never worse below $\delta = 0.3$ (dashed) but clearly worse above 0.5. (c) Drift grows with cache age (median, solid; 90th percentile, dashed), so the age limit $T$ rarely binds.}
  \label{fig:drift_diagnostic}
\end{figure}

\subsection{Problem Formulation}
\label{sec:formulation}
At each control step, a VLA model processes a set of visual tokens $\mathcal{V}$ together with the instruction tokens. We partition $\mathcal{V}$ into three disjoint sets: $\mathcal{R}$, the tokens we recompute; $\mathcal{U}$, the tokens we reuse from the cache; and $\mathcal{P}$, the tokens we prune. Prune-only and reuse-only methods are the two extremes of this partition: prune-only methods leave $\mathcal{U}$ empty, and reuse-only methods leave $\mathcal{P}$ empty. \method\ covers everything in between by making two decisions at each step: which tokens to place in $\mathcal{R}$, and, for every token outside $\mathcal{R}$, whether to reuse or prune it. This section addresses only the second decision. Taking $\mathcal{R}$ as given, we ask which signal determines whether reusing a token keeps the action closer to full recomputation than pruning it, and show that drift is such a signal. Section~\ref{sec:method} then describes how \method\ chooses $\mathcal{R}$ from importance and cache validity.

This decision depends on how far a token's cached state has moved from what the model would compute for the current frame. We measure this with \textbf{drift}, the relative change in a token's attention value at a shallow layer $\ell_s$ since its last recomputation:
\begin{equation}
    d_i = \frac{1}{H}\sum_{h=1}^{H}
    \frac{\lVert v_{i,h} - \bar{v}_{i,h} \rVert_2}{\lVert \bar{v}_{i,h} \rVert_2},
    \label{eq:drift}
\end{equation}
where $v_{i,h}$ is the current layer-$\ell_s$ value of token $i$ in head $h$, and $\bar{v}_{i,h}$ is the value stored at its last recomputation. We use $\ell_s = 2$, but any shallow layer works once $\delta$ is calibrated for it (Section~\ref{sec:sensitivity}). Because the layers up to $\ell_s$ run on every token at every step, drift comes at no extra cost.

\subsection{Token-level Diagnostic}
\label{sec:token_diag}
Intuitively, a cached KV with small drift stays close to what the model would compute for the current frame, so reusing it retains context that pruning would discard. A cached KV with large drift instead injects incorrect information into other tokens through attention, and the error compounds across layers. This suggests a drift threshold beyond which reuse becomes worse than pruning. However, intuition alone cannot tell whether such a threshold exists in practice, where it lies, or whether it depends on token importance or cache age. We answer these questions empirically with a controlled comparison: we take a group of tokens, handle them once by reuse and once by pruning while keeping everything else identical, and check which choice yields an action closer to full recomputation.

\textit{Setup.} We run full recomputation on LIBERO-Spatial (20 episodes per backbone; for OFT, one action is executed per query) and use its actions as the reference. Each comparison, which we call a probe, is defined by two choices. The first is the cache age $a$, the number of queries since the cache was computed: we build a cache holding the KV of every visual token in layers 3--31 from $a$ queries earlier ($a \in \{1, 2, 5, 10, 20, 40\}$; $a \le 20$ for OFT). The second is the group of probed tokens: we divide the visual tokens into three equal-sized levels of drift, measured at $\ell_s = 2$, and three equal-sized levels of importance, the attention a token receives from the decoding tokens (Section~\ref{sec:importance}), and sample 16 tokens from one of the nine resulting combinations, so that the probes cover every level of both. All other tokens are recomputed, and the 16 tokens are either reused from the cache or pruned in layers 3--31. We then compute
\begin{equation}
    \Delta = \lVert \mathbf{a}^{\mathrm{reuse}} - \mathbf{a}^{\mathrm{full}} \rVert_1
            - \lVert \mathbf{a}^{\mathrm{prune}} - \mathbf{a}^{\mathrm{full}} \rVert_1,
    \label{eq:reuse_prune}
\end{equation}
where $\mathbf{a}$ is the 7-dimensional action for OpenVLA and the first action of the chunk for OFT. A negative $\Delta$ means that reuse keeps the action closer to full recomputation than pruning does. We collect 14{,}442 probes on OpenVLA and 9{,}925 on OFT. Fig.~\ref{fig:drift_diagnostic}a,b plots $\Delta$ against the drift of the probed tokens, both pooled over all cache ages and separately for each age (cells with fewer than 30 probes or 5 episodes omitted).

\textit{When drift is small, reuse is no worse than pruning.} Below a drift of 0.3, the pooled $\Delta$ is negative on both backbones, with its 95\% confidence interval entirely below zero, and no individual cache age rises above zero. That age does not move this threshold is expected once drift is controlled for: an old cache is harmful mainly because its token has changed, which drift already measures, and drift grows with age (Fig.~\ref{fig:drift_diagnostic}c). Age still affects how much reuse helps: on OpenVLA, reuse remains better even for caches computed 20 or 40 queries earlier, whereas on OFT its advantage vanishes from age 10, although reuse never becomes harmful.

\textit{When drift is large, reuse is worse than pruning.} Above a drift of 0.5, the trend reverses on both backbones: reusing the stale KV yields 1.6$\times$ (OpenVLA) and 4.4$\times$ (OFT) the action error of pruning it. The harm grows with cache age: caches computed one query earlier stay close to zero even at this drift, whereas older caches are clearly worse than pruning.

Together, these observations show that a drift threshold separates tokens worth reusing from tokens better pruned, but its location depends on the backbone: between drifts of 0.3 and 0.5, reuse remains harmless on OpenVLA but is slightly worse than pruning on OFT. In principle, the threshold should therefore be calibrated per backbone with this diagnostic, which needs only full-recomputation rollouts and no training. For simplicity, we use a single value for both backbones, $\delta = 0.3$, the largest tested value below which reuse was never harmful on either; it is conservative for OpenVLA and matches OFT. End-to-end success is also insensitive to this choice: setting $\delta$ to 0.2 or 0.4 leaves the success rate statistically unchanged (Appendix~\ref{sec:sensitivity}).

\begin{figure*}[t]
  \centering
  \includegraphics[width=\textwidth]{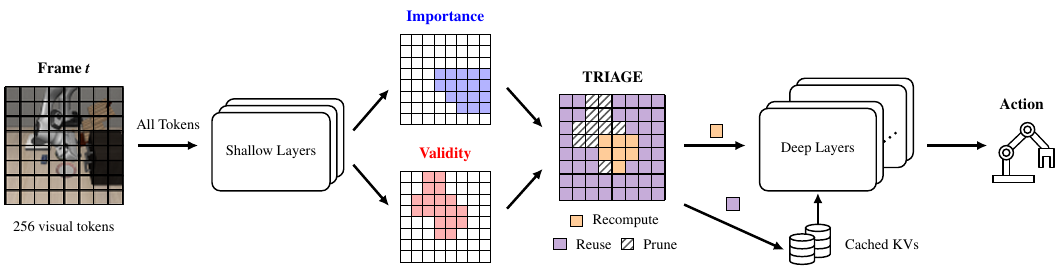}
  \caption{\textbf{Overview of \method.} Shallow layers mark each token as important (blue) and its cache as invalid (red). The budget $B$ goes to important, invalid tokens first; every other token is reused if valid and pruned otherwise, so $B$ alone sets the prefill latency.}
  \label{fig:overview}
\end{figure*}

\section{TRIAGE: Recompute, Reuse, or Prune}
\label{sec:method}

\subsection{Overview}
\label{sec:overview}
The layers up to $\ell_s$ run on all visual tokens and provide two signals per token: attention, which measures importance, and drift, which together with cache age determines validity (Fig.~\ref{fig:overview}). From these, \method\ assigns each token to recomputation, reuse, or pruning. The deep layers compute only the recomputed tokens and read the KVs of reused tokens from the cache.

\subsection{Importance}
\label{sec:importance}
The shallow signal $a_i$ is the attention that the decoding tokens assign to token $i$ at layer $\ell_s$, averaged over heads. Shallow attention is broad, whereas deep attention concentrates on task-critical regions, so tokens that look unimportant at a shallow layer can become important in deep layers~\cite{safe}.

To compensate, we adopt the deep-layer saliency forecast of SAFE-Pruner~\cite{safe}. At the first query, we record a memory $m$, the mean of the layer-0 and layer-$(L-1)$ attention to each visual token, as in SAFE-Pruner's official configuration. At every later query, $m$ is transferred to the current frame through patch correspondence: for each current token $i$, we find the first-frame token $\pi(i)$ whose projector output has the highest cosine similarity to that of $i$, and set $m'_i = m_{\pi(i)}$. Importance therefore follows an object even as it moves.

Unlike SAFE-Pruner, we never refresh $m$. SAFE-Pruner re-records $m$ at keyframes, selected when shallow attention departs from the forecast, and runs those queries with all tokens, which raises tail latency (Section~\ref{sec:main}). Since the purpose of a fixed budget is to bound the latency of every step, \method\ runs no full-token query at any step. A fixed $m$ remains usable because patch correspondence carries it to the current frame as objects move, and the current shallow attention $a_i$ still enters every importance score.

The importance score is $I_i = a_i + \lambda m'_i$ with $\lambda = 1$, as in SAFE-Pruner, which ranks tokens identically to fusing the two with equal weights. The important set $\mathcal{K}_t$ consists of the top $B + \kappa$ tokens by $I_i$, with $\kappa = 70$. We deliberately make $\mathcal{K}_t$ larger than the budget $B$ so that important tokens whose caches are still valid can be reused instead of recomputed.

\subsection{Validity}
\label{sec:validity}
The cache of token $i$ is valid if and only if
\begin{equation}
\mathrm{valid}_i = (\text{$i$ is cached}) \wedge (d_i < \delta) \wedge (\mathrm{age}_i \le T),
\label{eq:valid}
\end{equation}
where $d_i$ is the drift of Eq.~\eqref{eq:drift} and $\mathrm{age}_i$ is the number of queries since token $i$ was last recomputed.

\textit{Existence.} A token that has never been computed has no cache to reuse.

\textit{Drift.} We require $d_i < \delta$ with $\delta = 0.3$, the threshold calibrated in Section~\ref{sec:token_diag}; other backbones or tasks may call for a different value, which the same diagnostic provides. Selecting tokens for recomputation by such a change measure follows Eventful Transformers~\cite{eventful} and CacheBlend~\cite{cacheblend}. Because the reference value $\bar{v}_{i,h}$ is anchored at the last recomputation rather than at the previous query, slow changes accumulate and are eventually detected.

\textit{Age.} As a conservative safeguard on top of the drift test, we also require $\mathrm{age}_i \le T$ with $T = 10$. Drift is measured only at layer~$\ell_s$, and for tokens just below the threshold, the error of their deep-layer caches still grows with age (Appendix~\ref{sec:deep_error}). The limit rarely binds, because drift itself grows with age (Fig.~\ref{fig:drift_diagnostic}c), and removing it does not significantly change the success rate (Appendix~\ref{sec:sensitivity}).

\subsection{TRIAGE}
\label{sec:triage}
\begin{center}\footnotesize\setlength{\tabcolsep}{5pt}
\begin{tabular}{lll}
\toprule
 & cache valid & cache invalid\\
\midrule
important & reuse, or recompute by drift & \textbf{recompute first}\\
not important & reuse (free) & prune\\
\bottomrule
\end{tabular}
\end{center}

Given the budget $B$, the importance set $\mathcal{K}_t$, and the validity flags, each step partitions the visual tokens into three sets: recompute ($\mathcal{R}$), reuse ($\mathcal{U}$), and prune ($\mathcal{P}$). The budget $B$ is the knob that sets latency: since $|\mathcal{R}| = B$ at every step, the prefill latency is determined by $B$, so in deployment $B$ is chosen as the largest value that meets the latency target of the control loop. In our experiments, we set $B$ to match SAFE-Pruner's FLOPs (Section~\ref{sec:setup})
and sweep it in Section~\ref{sec:necessity}. The recompute set $\mathcal{R}$ is filled with $B$ tokens chosen in the following order of priority:

\begin{enumerate}
  \item[(i)] \textbf{Maintenance.} The $r=\mathrm{round}(\mu B)$ oldest cached
  tokens outside $\mathcal{K}_t$ are refreshed first, where $\mu$ is the
  fraction of the budget reserved for maintenance.
  \item[(ii)] \textbf{Mandatory recomputation.} Important tokens that cannot
  be reused are added in decreasing order of $I_i$. These are tokens in
  $\mathcal{K}_t$ whose cache is invalid, and tokens that have just entered
  $\mathcal{K}_t$, whose cache was computed while they were judged
  unimportant and therefore cannot be trusted as is.
  \item[(iii)] \textbf{Optional refresh.} The remaining slots go to important
  tokens with a valid cache, fastest-changing first, ranked by $a_i h_i$,
  where $h_i=\lVert x_i^t-x_i^{t-1}\rVert/\lVert x_i^t\rVert$ is the change of
  the layer-$(\ell_s-1)$ hidden state since the previous query.
\end{enumerate}
Every token not in $\mathcal{R}$ has its fate decided by validity alone,
independent of importance: it is reused if its cache is valid and pruned
otherwise, i.e., $\mathcal{U}=\{i\notin\mathcal{R}:\mathrm{valid}_i\}$ and
$\mathcal{P}=V\setminus(\mathcal{R}\cup\mathcal{U})$.

The three tiers play distinct roles. Tier~(ii) handles the one case that
cannot be resolved without computation: an important token whose cache is
invalid. Tier~(iii) keeps the caches of fast-changing important tokens up to
date. Tier~(i) keeps background reuse alive, i.e., the bottom-left cell of the
table above populated. Without it, a token outside $\mathcal{K}_t$ is never
recomputed; its cache expires at age $T$, and background reuse vanishes after
about $T$ queries. Since maintenance refreshes only cached tokens, a token
left uncached by the first query is computed only once it enters
$\mathcal{K}_t$.

\section{EXPERIMENTS}
\label{sec:exp}

\begin{figure}[t]
  \centering
  \includegraphics[width=\linewidth]{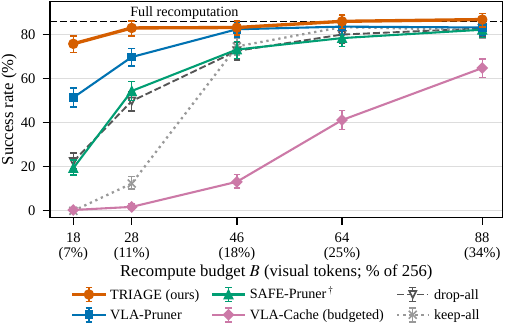}
  \caption{\textbf{Only combining pruning and reuse survives small budgets.} LIBERO-Object, OpenVLA (500 episodes per budget; Wilson 95\% intervals). x-axis: visual tokens recomputed per query, out of 256. Drop-all / keep-all prune / reuse every non-recomputed token. Drop-all degrades steadily and keep-all collapses below $B = 46$, while \method\ leads VLA-Pruner by 24.4\pp\ at $B = 18$.}
  \label{fig:teaser}
\end{figure}

\subsection{Setup}
\label{sec:setup}
\textit{Benchmarks.} We evaluate on the four LIBERO suites~\cite{libero} (500 episodes each) and on SimplerEnv~\cite{simpler} Google Robot (Move Near under Visual Matching; Move Near and Pick Coke Can under Variant Aggregation), keeping only the tasks on which full recomputation is clearly above the floor.

\textit{Backbones.} OpenVLA-7B~\cite{openvla} uses 256 visual tokens and decodes actions autoregressively; OpenVLA-OFT~\cite{oft} uses 512 visual tokens from two cameras and predicts an 8-step action chunk in a single forward pass. We use their public LIBERO checkpoints, and the base OpenVLA checkpoint on SimplerEnv.

\textit{Baselines.} We compare against full recomputation and four training-free methods at their default settings: VLA-Cache~\cite{vlacache}, VLA-Pruner~\cite{vlapruner}, SAFE-Pruner~\cite{safe}, and SpecPrune-VLA~\cite{specprune}. For SAFE-Pruner on OpenVLA and for SpecPrune-VLA, we use our own implementations.

\textit{Implementation.} \method\ recomputes $B = 88$ visual tokens on OpenVLA and 163 on OFT (91 on OFT-Goal), matching SAFE-Pruner's FLOPs, with $T = 10$ and $\mu = 0.06$. All hyperparameters were tuned on development episodes, mainly from LIBERO-Spatial, and frozen before evaluation.

\textit{Metrics.} We report success rate, analytic prefill FLOPs, and latency on a Jetson AGX Orin and, for reference, an RTX 3090 (both bf16): prefill time, end-to-end time per policy call, and p99, over 250 queries per suite that replay the same recorded observation stream.

\subsection{Both Pruning and Reuse Are Needed}
\label{sec:necessity}
Fig.~\ref{fig:teaser} isolates how the tokens that are not recomputed should be handled. We compare \method\ with two variants that share its importance ranking and recompute the same $B$ tokens: \emph{drop-all} prunes all remaining tokens, and \emph{keep-all} reuses all of them. We also cap VLA-Cache's recomputed set at $B$, and sweep $B$ from 88 down to 18 (34\% to 7\% of the visual tokens) on LIBERO-Object with OpenVLA. Comparisons with VLA-Pruner use a paired bootstrap over episodes.

The two extremes fail for different reasons. Drop-all degrades steadily as the budget shrinks, reaching 22.4\% at $B = 18$: the recomputed set no longer covers every task-relevant region, and the missed tokens are discarded even when their caches are still accurate. SAFE-Pruner, which also prunes every token it does not keep, follows the same trend (19.4\%). Keep-all instead stays competitive at large budgets but collapses below $B = 46$ (0.0\% at $B = 18$), as stale entries pass incorrect context to the recomputed tokens through attention; budgeted VLA-Cache likewise falls to 0.4\%. As a result, neither extreme is best across budgets: keep-all is ahead of drop-all at $B = 46$ and 64, and drop-all is ahead at $B = 28$.

\method\ combines both. It matches full recomputation at $B = 64$ and stays within 3\pp\ of it down to $B = 28$; at $B = 18$, it still reaches 75.8\%, while every other method is at or below 51.4\%. It is statistically indistinguishable from VLA-Pruner, the strongest baseline, at $B \ge 46$, but leads it by 13.2\pp\ at $B = 28$ and 24.4\pp\ at $B = 18$.

\begin{figure*}[t]
  \centering
  \includegraphics[width=\textwidth]{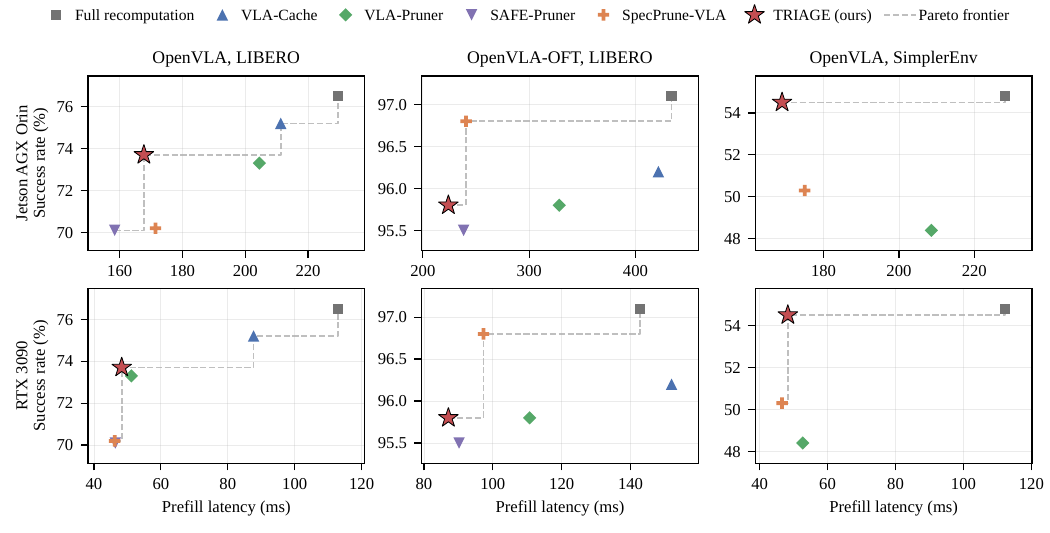}
  \caption{\textbf{\method\ lies on the Pareto frontier in every setting.} Success rate versus prefill latency on a Jetson AGX Orin (top) and an RTX 3090 (bottom), averaged over the LIBERO suites or the SimplerEnv settings; dashed: Pareto frontier. No baseline is both more accurate and faster.}
  \label{fig:tradeoff}
\end{figure*}

\begin{table*}[t]
\caption{Results on LIBERO: \method\ stays within 3\pp\ of full recomputation with the lowest Jetson p99 latency}
\label{tab:libero}
\centering\footnotesize\renewcommand{\arraystretch}{0.95}\setlength{\tabcolsep}{4.5pt}
\begin{threeparttable}
\begin{tabular}{lcccccccccccc}
\toprule
 & & & & & & & \multicolumn{6}{c}{Latency (ms)}\\
\cmidrule(lr){8-13}
 & & & & & & & \multicolumn{3}{c}{Jetson AGX Orin} & \multicolumn{3}{c}{RTX 3090}\\
\cmidrule(lr){8-10}\cmidrule(lr){11-13}
Method & Spatial & Object & Goal & Long & Avg & FLOPs & Prefill & E2E & p99 & Prefill & E2E & p99\\
\midrule
\multicolumn{13}{l}{\textit{OpenVLA}}\\
Full recomputation & 85.0 & 86.0 & 79.2 & 56.0 & 76.5 & 1.00$\times$ & 229.5 & 1261.3 & 1307.1 & 112.9 & 281.4 & 286.3\\
VLA-Cache~\cite{vlacache} & \textbf{84.8} & \textbf{87.2} & \textbf{78.2} & \textbf{50.4} & \textbf{75.2} & 0.78$\times$ & 211.3 & 1365.7 & 1415.3 & 87.7 & 278.7 & 289.5\\
VLA-Pruner~\cite{vlapruner} & \underline{84.4} & 83.6 & 76.2 & \underline{49.2} & 73.3 & \textbf{0.39}$\times$ & 204.5 & 1368.7 & 1428.5 & 51.2 & 246.1 & 315.9\\
SAFE-Pruner$^\dagger$~\cite{safe} & 81.4 & 82.2 & 73.4 & 43.2 & 70.1 & 0.45$\times$ & \textbf{158.4} & \textbf{1166.6} & 1255.9 & \underline{46.4} & \textbf{238.7} & 308.9\\
SpecPrune-VLA$^\ddagger$~\cite{specprune} & 82.4 & 81.4 & 73.6 & 43.2 & 70.2 & \underline{0.40}$\times$ & 171.4 & 1177.0 & \underline{1248.1} & \textbf{46.2} & \textbf{238.7} & \textbf{242.6}\\
\textbf{\method\ (ours)} & 81.4 & \underline{86.8} & \underline{77.4} & \underline{49.2} & \underline{73.7} & 0.45$\times$ & \underline{167.7} & \underline{1173.4} & \textbf{1228.2} & 48.3 & \underline{241.8} & \underline{245.9}\\
\midrule
\multicolumn{13}{l}{\textit{OpenVLA-OFT}}\\
Full recomputation & 98.8 & 97.6 & 98.4 & 93.6 & 97.1 & 1.00$\times$ & 433.9 & 653.0 & 674.8 & 142.7 & 181.0 & 184.3\\
VLA-Cache~\cite{vlacache} & 97.0 & 96.8 & \underline{97.8} & \textbf{93.4} & \underline{96.2} & 0.77$\times$ & 421.5 & 702.3 & 774.2 & 151.9 & 202.4 & 206.2\\
VLA-Pruner~\cite{vlapruner} & 97.4 & 97.4 & \underline{97.8} & 90.8 & 95.8 & 0.53$\times$ & 328.3 & 555.2 & 765.8 & 110.7 & 152.2 & 200.6\\
SpecPrune-VLA$^\ddagger$~\cite{specprune} & \textbf{99.2} & \underline{98.2} & \textbf{98.0} & \underline{91.6} & \textbf{96.8} & \underline{0.46}$\times$ & 240.6 & \underline{460.7} & \underline{672.4} & 97.3 & 136.0 & \underline{183.7}\\
SAFE-Pruner~\cite{safe} & 97.8 & 97.4 & 95.6 & 91.4 & 95.5 & \textbf{0.44}$\times$ & \underline{238.3} & 466.5 & 768.0 & \underline{90.2} & \underline{130.9} & 200.4\\
\textbf{\method\ (ours)} & \underline{98.6} & \textbf{98.4} & \underline{97.8} & 88.6 & 95.8 & \textbf{0.44}$\times$ & \textbf{224.0} & \textbf{443.1} & \textbf{461.1} & \textbf{87.1} & \textbf{126.5} & \textbf{138.4}\\
\bottomrule
\end{tabular}
\begin{tablenotes}[flushleft]\scriptsize
\item Success rate (\%) over 500 episodes per suite. FLOPs: mean prefill FLOPs per query relative to full recomputation. Latency in bf16 on a Jetson AGX Orin and an RTX 3090, replaying the same observation stream; E2E: one policy call; p99: E2E p99. On OpenVLA-OFT, one query predicts an 8-action chunk. Bold: best; underline: second best among the accelerated methods within each backbone (full recomputation excluded).
\item[$\dagger$] Our port to OpenVLA (no official code); recomputes the same 88 visual tokens as \method.
\item[$\ddagger$] Our reimplementation of the static stage only (no public code). 
\end{tablenotes}
\end{threeparttable}
\end{table*}

\begin{table*}[t]
\caption{OpenVLA on SimplerEnv (Google Robot): without retuning, \method\ matches full recomputation}
\label{tab:simpler}
\centering\footnotesize\renewcommand{\arraystretch}{0.95}\setlength{\tabcolsep}{4.5pt}
\begin{threeparttable}
\begin{tabular}{lcccccccccc}
\toprule
 & \multicolumn{2}{c}{Move Near} & Coke Can & & \multicolumn{6}{c}{Latency (ms)}\\
\cmidrule(lr){2-3}\cmidrule(lr){4-4}\cmidrule(lr){6-11}
 & & & & & \multicolumn{3}{c}{Jetson AGX Orin} & \multicolumn{3}{c}{RTX 3090}\\
\cmidrule(lr){6-8}\cmidrule(lr){9-11}
Method & VM & VA & VA & Avg & Prefill & E2E & p99 & Prefill & E2E & p99\\
\midrule
Full recomputation & 50.0 & 60.4 & 53.9 & 54.8 & 228.2 & 1230.6 & 1263.4 & 112.3 & 275.4 & 278.7\\
\midrule
VLA-Pruner~\cite{vlapruner} & 40.8 & 53.8 & \underline{50.5} & 48.4 & 208.6 & 1385.0 & 1447.2 & 52.7 & 242.6 & 312.2\\
SpecPrune-VLA$^\ddagger$~\cite{specprune} & \underline{45.4} & \underline{56.7} & 48.8 & \underline{50.3} & \underline{175.0} & \underline{1189.2} & \underline{1254.9} & \textbf{46.6} & \textbf{236.0} & \underline{305.7}\\
\textbf{\method\ (ours)} & \textbf{50.4} & \textbf{61.2} & \textbf{51.9} & \textbf{54.5} & \textbf{169.0} & \textbf{1179.4} & \textbf{1224.2} & \underline{48.3} & \underline{238.6} & \textbf{279.5}\\
\bottomrule
\end{tabular}
\begin{tablenotes}[flushleft]\scriptsize
\item Success rate (\%) over 240 episodes per Move Near setting and 297 for Coke Can; VM: Visual Matching, VA: Variant Aggregation. Latency and highlighting as in Table~\ref{tab:libero}.
\item[$\ddagger$] As in Table~\ref{tab:libero}.
\end{tablenotes}
\end{threeparttable}
\end{table*}

\subsection{Main Results}
\label{sec:main}
\method\ offers the best trade-off between success rate and latency. On every backbone, benchmark, and device, it lies on the Pareto frontier of success rate and prefill latency (Fig.~\ref{fig:tradeoff}) and stays within 3\pp\ of full recomputation in average success rate. Baselines that match its success rate are slower, and those that are as fast or faster are less accurate. On the Jetson, it also has the lowest p99 latency in every setting (Table~\ref{tab:libero}--\ref{tab:simpler}).

\textit{OpenVLA.} \method\ reaches 73.7\%, 3.6\pp\ above SAFE-Pruner at the same FLOPs and 3.5\pp\ above SpecPrune-VLA, the two baselines with similar latency; the gain is largest on Long (+6.0\pp). Only VLA-Cache (75.2\%) and full recomputation are more accurate, and both are slower. VLA-Pruner reaches a comparable 73.3\% with fewer FLOPs, but on the Jetson this reduction does not carry over: its end-to-end step is 17\% longer than that of \method, as is VLA-Cache's (16\%), and both are slower than even full recomputation. \method\ reduces prefill latency to 0.73$\times$ that of full recomputation on the Jetson and to 0.43$\times$ on the RTX 3090. The end-to-end gain on the Jetson is smaller (0.93$\times$) because 82\% of the full-recomputation step lies outside prefill.

\textit{OpenVLA-OFT.} \method\ averages 95.8\%, 1.3\pp\ below full recomputation and 0.3\pp\ above SAFE-Pruner at the same FLOPs. SpecPrune-VLA is 1.0\pp\ more accurate but slower: its prefill is 7\% longer than that of \method, and its p99 latency is 1.46$\times$ higher on the Jetson. \method\ has the lowest prefill latency on both devices (0.52$\times$ that of full recomputation on the Jetson) and the lowest end-to-end latency on the Jetson (0.68$\times$). Because it recomputes a fixed budget at every query, its p99 latency is only 4\% above its mean, whereas SAFE-Pruner's reaches 1.6$\times$ its mean because its keyframes run full-token inference~\cite{safe}.

\textit{SimplerEnv.} The same settings transfer without retuning (Table~\ref{tab:simpler}). \method\ averages 54.5\% against 54.8\% for full recomputation, while SpecPrune-VLA and VLA-Pruner trail it by 4.2 and 6.1\pp. On the Jetson, it again has the lowest prefill (0.74$\times$ that of full recomputation), end-to-end, and p99 latency.

\section{DISCUSSION AND LIMITATIONS}
\label{sec:discussion}

\textbf{The hyperparameter tuning.} \method\ is based on several hyperparameters, including $\delta$, $T$, $\kappa$, $\lambda$, and $\mu$. We tuned them on development episodes and froze them before evaluation, but their values are set empirically rather than derived from first principles.

\textbf{Workload-dependent thresholds.} The drift threshold $\delta$ that separates reuse from pruning is not fixed. In our diagnostic, reuse became harmful from a drift of 0.5 on OpenVLA but from 0.3 on OpenVLA-OFT. This point may shift further with the backbone, the task, and the speed of the scene. Therefore, an adaptive method that adjusts the threshold during deployment is needed.

\textbf{Exploiting motion.} \method\ detects change only after it happens, through drift. Optical flow, as used by LAC~\cite{lac}, motion vectors produced by video codecs, or a learned patch selector such as AutoGaze~\cite{autogaze} could separate static regions from moving ones before the backbone runs. Reusing static regions more aggressively and recomputing moving regions first could improve latency and accuracy at the same time.

\section{CONCLUSION}
\label{sec:conclusion}
This paper addressed how to handle the visual tokens that a VLA model does not recompute. Previous work has given this question a fixed answer: prune them all or reuse them all. Our token-level diagnostic shows that the right answer depends on drift: tokens with accurate caches are better reused, whereas stale ones are better pruned. Based on this observation, \method\ decides for each token whether to recompute, reuse, or prune it. As a result, it offers the best trade-off between success rate and latency among the methods we compared: it stays within 3\pp\ of full recomputation, no baseline is both more accurate and faster, and its advantage grows as the recomputation budget shrinks. The drift threshold may need recalibration for other backbones, but the same diagnostic procedure provides it. These results show that efficient VLA inference requires both pruning and reuse, not one or the other.

\section*{APPENDIX}
\subsection{Sensitivity Analysis}
\label{sec:sensitivity}
\begin{figure}[t]
  \centering
  \includegraphics[width=\columnwidth]{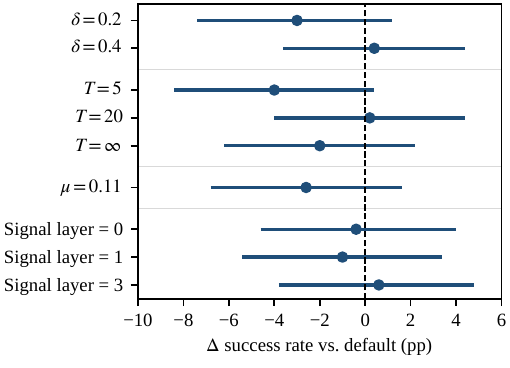}
  \caption{\textbf{\method\ is robust to its hyperparameters.} LIBERO-Goal, OpenVLA, $B = 28$ (default: 70.6\%). Each row changes one setting from the default ($\delta = 0.3$, $T = 10$, $\mu = 0.06$, signal layer 2); paired difference with 95\% bootstrap interval over 500 episodes. Every interval contains zero.}
  \label{fig:sensitivity}
\end{figure}

We vary one hyperparameter at a time in LIBERO-Goal, a suite not used for tuning, with OpenVLA at $B = 28$ (Fig.~\ref{fig:sensitivity}). No change moves the success rate by more than 4\pp\ from the default (70.6\%), and every 95\% interval contains zero. Tightening $\delta$ or $T$ costs the most, as it prunes caches that could have been reused. Removing the age limit altogether ($T = \infty$) lowers the success rate from 70.6\% to 68.6\% (paired difference $-2.0$\pp, 95\% CI $[-6.2, +2.2]$), and no tested value of $T \in \{5, 10, 20, \infty\}$ changes it significantly. The drift test thus accounts for most of the filtering of stale caches, and $T$ acts as a safeguard.

\subsection{Deep-Layer Error of Reused Caches}
\label{sec:deep_error}
On the OpenVLA probes of Section~\ref{sec:token_diag}, we measure the relative change in the attention the 16 probed tokens receive at layers 15 and 31 when they are reused instead of recomputed, with drift fixed to a bin so that age and drift are not confounded (Table~\ref{tab:deep_error}). Layer~31 deviates more than layer~15 in every cell. The deviation does not grow with age for drift below 0.1, but grows from 0.17 to 0.27 at layer~31 between ages 1 and 10 for drift between 0.2 and 0.3; the age limit $T$ targets these tokens.

\begin{table}[t]
\caption{Error of reused caches grows with age only just below $\delta$ (OpenVLA, median relative change in attention)}
\label{tab:deep_error}
\centering\footnotesize\setlength{\tabcolsep}{4pt}
\begin{threeparttable}
\begin{tabular}{lcccc}
\toprule
Drift & Age 1 & Age 5 & Age 10 & Age 20\\
\midrule
\multicolumn{5}{l}{\textit{Layer 15}}\\
$[0, 0.1)$   & 0.04 & 0.03 & 0.02 & 0.03\\
$[0.1, 0.2)$ & 0.06 & 0.10 & 0.09 & 0.09\\
$[0.2, 0.3)$ & 0.11 & 0.14 & 0.19 & 0.21\\
\midrule
\multicolumn{5}{l}{\textit{Layer 31}}\\
$[0, 0.1)$   & 0.08 & 0.08 & 0.06 & 0.07\\
$[0.1, 0.2)$ & 0.12 & 0.18 & 0.15 & 0.18\\
$[0.2, 0.3)$ & 0.17 & 0.21 & 0.27 & 0.28\\
\bottomrule
\end{tabular}
\begin{tablenotes}[flushleft]\scriptsize
\item Layer 31, drift $[0.2, 0.3)$: 95\% CI [0.15, 0.20] at age 1 and [0.24, 0.29] at age 10.
\end{tablenotes}
\end{threeparttable}
\end{table}

\section*{ACKNOWLEDGMENT}
The authors used Claude for proofreading and coding assistance and verified all outputs.


\end{document}